\documentclass{article}

\usepackage{PRIMEarxiv}
\usepackage[utf8]{inputenc} % allow utf-8 input
\usepackage[T1]{fontenc}    % use 8-bit T1 fonts
\usepackage{hyperref}       % hyperlinks
\usepackage{url}            % simple URL typesetting
\usepackage{booktabs}       % professional-quality tables
\usepackage{amsfonts}       % blackboard math symbols
\usepackage{nicefrac}       % compact symbols for 1/2, etc.
\usepackage{microtype}      % microtypography
\usepackage{lipsum}
\usepackage{fancyhdr}       % header
\usepackage{graphicx}       % graphics
\graphicspath{{media/}}     % organize your images and other figures under media/ folder
\usepackage{booktabs}      %三线表
\usepackage{multirow}
\usepackage{enumerate}
\title{A method for incremental discovery of financial event types based on anomaly detection}
\author{
  Dianyue Gu, Zixu Li, Zhenhai Guan, Rui Zhang*,Lan Huang \\
  Key Laboratory of Symbolic Computation and Knowledge Engineering of MOE \\
  Jilin University \\
  Changchun,China\\
  \texttt{\{gudy21, zixu22, zhguan22\}@mails.jlu.edu.cn,\{rui,huanglan\}@jlu.edu.cn} \\
}

\begin{document}

\maketitle

\begin{abstract}
Event datasets in the financial domain are often constructed based on actual application scenarios, and their event types are weakly reusable due to scenario constraints; at the same time, the massive and diverse new financial big data cannot be limited to the event types defined for specific scenarios. This limitation of a small number of event types does not meet our research needs for more complex tasks such as the prediction of major financial events and the analysis of the ripple effects of financial events. In this paper, a three-stage approach is proposed to accomplish incremental discovery of event types. For an existing annotated financial event dataset, the three-stage approach consists of: for a set of financial event data with a mixture of original and unknown event types, a semi-supervised deep clustering model with anomaly detection is first applied to classify the data into normal and abnormal events, where abnormal events are events that do not belong to known types; then normal events are tagged with appropriate event types and abnormal events are reasonably clustered. Finally, a cluster keyword extraction method is used to recommend the type names of events for the new event clusters, thus incrementally discovering new event types. The proposed method is effective in the incremental discovery of new event types on real data sets.
\end{abstract}

\keywords{Financial event types \and Anomaly detection \and New type discovery \and Deep embedding clustering \and Semi-supervised learning}

\section{Preface}
The leap forward in internet technology has brought about massive growth in diverse financial data, of which the assessment, prediction, and analysis of the ripple effects of financial events is an important research area of national importance\cite{r1}. The widespread use of deep neural network methods has led to the construction of datasets in the financial sector, and datasets such as DuEE-fin have received widespread attention and recognition. However, most of the existing datasets are designed for specific tasks, which have the disadvantages of limited event types and poor generalization, making it difficult to meet our research needs for more complex tasks such as the prediction of major financial events and the analysis of the ripple effects of financial events. At the same time, the diversity and rapid growth of financial big data require the annotation of large amounts of new unlabeled data. On the one hand, there is a need for more accurate type annotation of large amounts of new data; on the other hand, there is a need to detect possible new types in new data and to reasonably discover and annotate new data types. Therefore, there is an urgent need for incremental new type discovery methods to correctly label the new financial events that are constantly emerging.

Event classification is a branch of text classification tasks. In recent years, many researchers have applied machine learning algorithms to event classification tasks, such as logistic regression classification\cite{r2}, parsimonious Bayesian classification\cite{r3}, and support vector machine classification.\cite{r4}The most important of these algorithms is Wang Yang\cite{r5} et al. They built an SVM-based classification model for very short texts, and the experimental results showed that the classification accuracy of this method reached 98\% by training and testing $9906$ very short texts. However, the machine learning-based text classification method has the disadvantages of high vector dimensionality and sparse data, which accelerate the disappearance of features\cite{r6}. Later, with the rise of deep learning, Xu et al\cite{r7} used a subset of the THUCNews data set to build a convolutional neural network model. The accuracy of classification prediction is 94.8\%, which is 11.7\% higher than the traditional text classification method.Although the above-mentioned classification methods have high accuracy, they are not suitable for practical applications in the face of the increasing number of new data samples, which are trained first and tested later. Incremental learning models that can predict while training and add new samples have emerged\cite{r8}. Zhao et al\cite{r9}proposed an incremental learning approach for text classification, which can improve the structure of the model and modify the loss function on top of the existing model to achieve incremental learning without forgetting the previously learned knowledge. However, incremental learning methods usually require that the nature of each type of sample is not too different, and when the nature and difficulty of the task are too different, the performance of most incremental learning methods will be severely degraded, even lower than that of traditional models. In addition, some studies have shown that most current incremental learning methods are sensitive to model structure, data properties, and super parameter settings, so it is interesting to explore incremental discovery methods that perform more robustly. In this paper, we propose a new type of incremental discovery method with anomaly detection, which can overcome the difficulty of large class differences, and experiments show that the method has certain stability and effectiveness.

The contribution of this paper is : 
\begin{enumerate}[1.]
\item A semi-supervised deep clustering model with anomaly detection was used to classify the data into normal and abnormal events, namely events that were not of known types, for a set of financial events with a mixture of original and unknown event types;
\item The normal events are tagged with the appropriate event type, namely the number of original data is expanded and the abnormal events are logically clustered into new event clusters;
\item The cluster keyword extraction method is used to suggest the type names of events for new event clusters, thus incrementally discovering new event types to improve the annotation effect.
\end{enumerate}

\section{Problem Definition}
We define the basic model for the discovery of new event types as:
\begin{equation}
M=E_\phi \left ( C_\omega\left ( F_\theta \left ( D^1 \right ),D^2  \right )   \right ) 
\end{equation}
It consists of a feature extractor $F_\theta \left ( \cdot \right )$  and anomaly detectors $C_\omega \left ( \cdot  \right )$ and a new type discoverer $E_\phi \left ( \cdot  \right )$  where $\theta$ ,$\omega$ and $\phi$  are the parameters of the feature extractor, anomaly detector, and new type discoverer, respectively,$D^1$ denotes a data set with known base class with annotations,$D^2$ denotes the dataset to be processed without annotation, Namely $D^{1} \cap D^{2}=\emptyset$,$D_{n}=\left\{\left(x_{i}, y_{i}\right)\right\}_{i=1}^{\left|D^{n}\right|}$ (${\left|D^{n}\right|}$ denotes the total number of samples,$n=1,2,...$),$x_i$ denotes the sample, and the corresponding category label is $y_i$, then $y_i\in D^1$ is known, $y_i\in D^2$ is unknown to be determined.

Denote the class set of the $n$th data set ($n=1,2,...$) by $T^n$, $T_{x_i}=y_i$ indicates that the type label of sample $x_i$ is $y_i$, for any dataset, their category sets can be intersecting or disintersecting, namely $T^{1} \cap T^{2}=\emptyset$ or $T^{1} \cap T^{2}\ne \emptyset$ , which correspond to the two cases of new event type discovery: when $T^{1} \cap T^{2}=\emptyset$, all the sample categories in the data to be processed are new to the base class, and each sample cannot be classified into any of the event types in the base class; when $T^{1} \cap T^{2}\ne \emptyset$, the data set to be processed contains samples that can be grouped into known base classes, so that the samples that can be grouped into base classes are expanded into known base classes at the same time as new types are discovered, thus simultaneously expanding the number of sample instances in the base class data set. In real-life tasks, more of the data sets to be processed have class intersections with the base class data sets, so this paper is concerned with the second case.

In this task, the following definitions are given $1-8$:
\begin{itemize}
\item Event: An event or change of state that occurs at a specific point in time or period, within a specific geographical area, involving one or more actors and consisting of one or more actions.
\item Known event: An instance of an event in a known base class dataset, $x_i\in D^1$.
\item Unknown events: New event instances in the pending dataset, $x_i\in D^2$.
\item Normal events: All events in the base class dataset and the data set to be processed whose event type belongs to T1, $\left ( \left ( x_i\in D^1\parallel x_i\in D^2 \right ) \cap T_{x_i}\in T^1 \right )$ .
\item Known normal events: Equivalent to known event,$x_i\in D^1\cap T_{x_i}\in T^1$.
\item Unknown normal events: events that do not appear in $D^1$ but the event type is in $T^1$, $x_i\in D^2\cap T_{x_i}\in T^1$.
\item Abnormal event: Events that neither appear in $D^1$ nor belong to any event type in $D^1$ .
\item New event type: The type of event to be found for the anomalous event.
\end{itemize}

Task 1 Anomaly Detection: We refer to the process of identifying abnormal events as the anomaly detection process.

Task 2 New event type discovery: We refer to the process of identifying the type of event to which a new event should belong as the process of new event type discovery.

$F_\theta \left ( D^1 \right )$  represents the clustering feature extracted by the feature extractor $F_\theta \left ( \cdot \right )$ on the known base class data set $D^1$, $C_\omega\left ( F_\theta \left ( D^1 \right ),D^2  \right )$ represents the abnormal event set identified by the anomaly detector $C_\omega\left ( \cdot  \right )$ on the basis of the clustering feature recognition result $C_\omega\left ( \cdot  \right )$ according to the clustering feature recognition result $F_\theta \left ( D^1 \right )$ on the basis of the anomaly detection of $D^2$, and $E_\phi\left ( C_\omega \left ( F_\theta \left (  D^1\right ) ,D^2 \right )  \right )$ represents the new event type discovery operation by the new type finder $E_\phi\left ( \cdot  \right )$ on the exception event set $C_\omega\left ( F_\theta \left ( D^1 \right ),D^2  \right )$ in $D^2$. Generally, the new type discovery problem includes the following three processes:
\begin{enumerate}[1)]
\item The model will first learn from the existing base class data set $D_{n}=\left\{\left(x_{i}, y_{i}\right)\right\}_{i=1}^{\left|D^{n}\right|}$ to obtain a feature extractor and base class classifier with good performance. At this stage, each base class (assuming $b=\left | T^1 \right |$ ) contains certain training samples.
\item The anomaly detection process uses a semi-supervised deep clustering model with anomaly detection to classify the data into normal and abnormal events, form reasonable clusters of abnormal events, and expand the unknown normal events into the base class data set.
\item We discover new event types based on the characteristics of clustered clusters by suggesting event type names for new event clusters through cluster keyword extraction methods.
\end{enumerate}

The goal of new event type discovery is: for a set of financial event data with mixed original event types and unknown event types, we learn the class characteristics of the existing data, first perform anomaly detection on the samples to be processed, identify the events that cannot be classified into the existing base class, and expand the unknown normal events that can be assigned to the existing base class into the data set, and finally cluster the abnormal events to discover $N$ new classes. This means that with the whole model, the $b+N$ classes of events can be correctly identified at the same time while expanding the number of instances in the base class dataset.
\begin{figure}
    \centering
    \includegraphics[width=16cm,height=12cm]{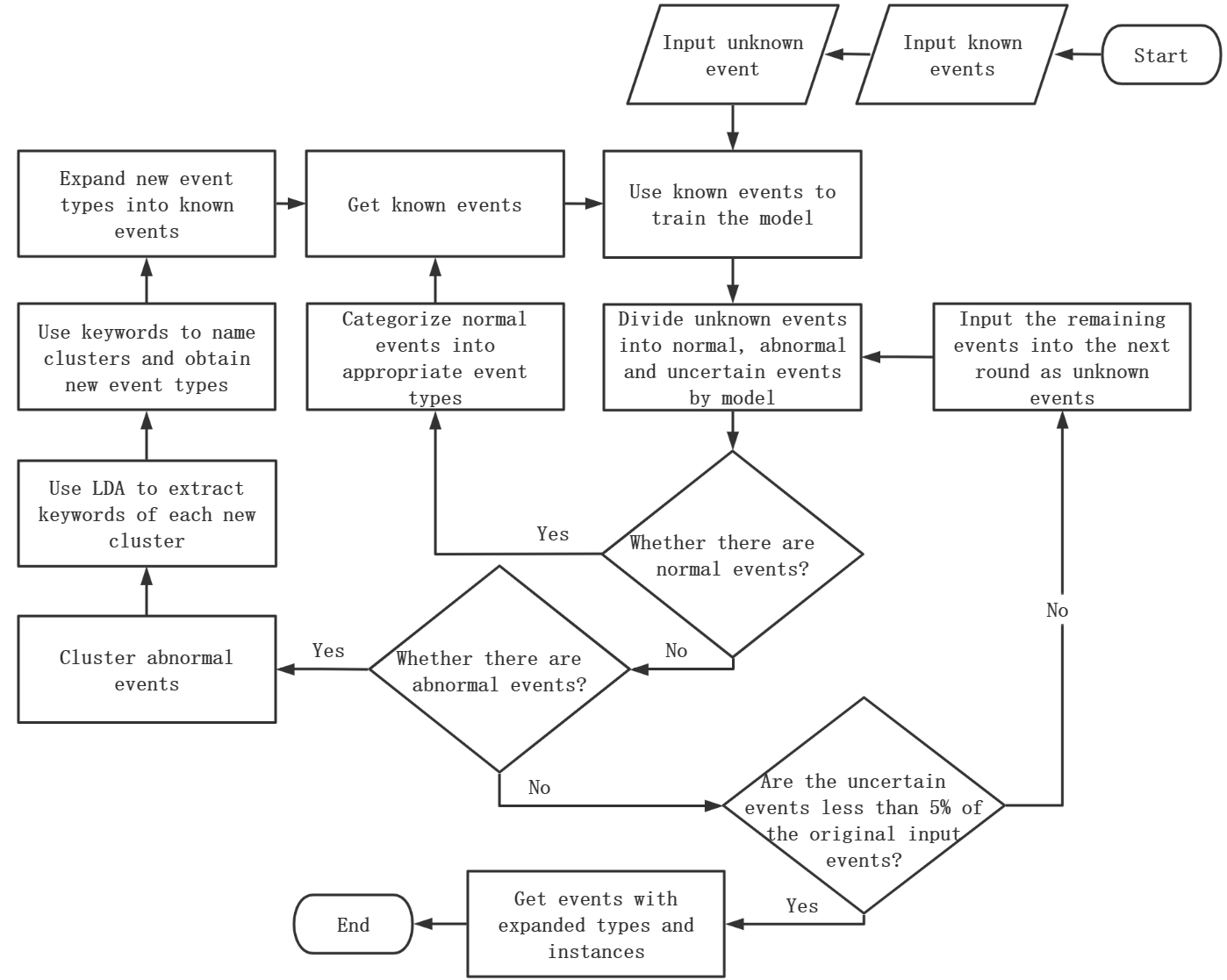}
    \caption{Flow chart of incremental discovery model}
    \label{fig:my_label}
\end{figure}

\section{Method}
\subsection{Base Class Dataset Expansion Methods for Joint Abnormal Event Detection}
The task of this module is to consider the problem of assigning a set of unknown events that do not exist in the base class dataset to a known type. First, In the data set$D^2$ to be processed , we separate the exception event set $E$ that cannot be described by any existing type in the base class data set. Next, we assign the remaining $n$ normal events $\left \{ x_i\in D^1 \right \} _{i=1}^{\left|D^{1}\right|}$to into $\left | C_1 \right |$ (the number of event types in the base class) clusters,  where each cluster is represented by a center of mass $\left \{ u_j \right \} _{j=1}^{\left|C^{1}\right|}$ and corresponds to an event type. This is a hard clustering task, where each event with a unique identifier is assigned only one event type.

We propose a semi-supervised deep embedding clustering model that jointly identifies abnormal events and assigns normal events to existing event types, namely the proposed incremental discovery model for financial event types based on anomaly detection (hereafter referred to as incremental discovery model). For n normal events embedded in the vector space $X$, the incremental discovery model uses a nonlinear transformation $f_\theta :X\to Z$ (where $\theta$ is the learnable parameter of the encoder) of the self-encoder to represent the semantic features of the normal events in the low-dimensional potential space $Z$. After embedding normal events into the potential space $Z$, the incremental discovery model clusters normal events (including unknown normal events) using the event type information present in the base class, which is a semi-supervised classification process. The incremental discovery model also learns the reconstruction of normal events from the potential space $X$ to its original vector space $X$ by minimizing the reconstruction error. Since no abnormal events are used in the training phase, the incremental discovery model can identify abnormal events with large reconstruction errors.

Figure 2 illustrates the incremental discovery model with a deep encoder-decoder structure. First, the encoder layer learns to embed the input $x$ into a low-dimensional potential representation $z$, and decoder layer learning to reconstruct $z$ back into the original embedding (top of Figure 2), and the reconstructed embedding is $\hat{x}$. The encoder layer is further trained to cluster the latent representations under the supervision of pairwise constraints $A$ (middle of Fig. 2), and the number of clusters is the number of event types in the base class. By minimizing the Kullback-Leibler scattered cluster loss (KL loss) and the constraint loss, the incremental discovery model learns to represent and cluster $x$ in the potential space.
\begin{figure}
    \centering
    \includegraphics[width=10cm,height=8cm]{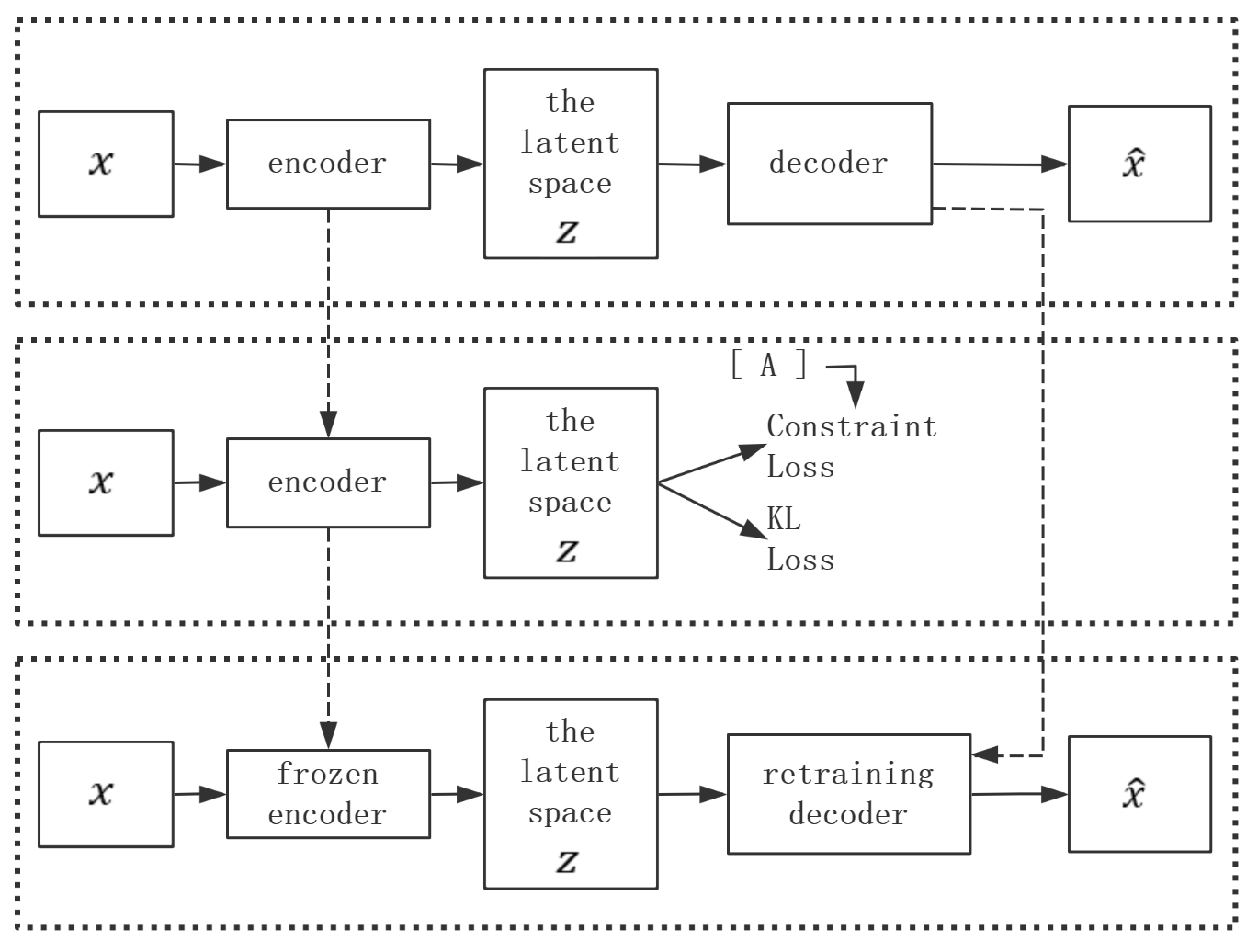}
    \caption{Framework of anomaly detection process}
    \label{fig:my_label}
\end{figure}

The encoder in the incremental discovery model receives the event text $\left \{ x_i\in X \right \} _{i=1}^{\left|n\right|}$as input and returns its feature representation $\left \{ z_i\in Z \right \} _{i=1}^{\left|n\right|}$ in the potential space $Z$ as the output. Then, we use supervised initialization to obtain the initial center of mass $\left \{ u_j \right \} _{j=1}^{\left|k\right|}$ in the space $Z$ where $k$ is the number of event types in the base class dataset. We first initialize the cores using information about the type labels of normal events and then start clustering normal events. We define each center of mass $\left \{ u_j \right \} _{j=1}^{\left|k\right|}$ as the average of a vector of normal events (in the potential space $Z$) that share the same event type. Note that we initialize the cores using only the events already in the base class.

After training the encoder of the incremental discovery model to embed and cluster normal events in the potential space $Z$, we freeze the encoder and train the decoder to map the events $\left \{ z_i\in Z \right \} _{i=1}^{\left|n\right|}$ back to its original representation $\left \{ x_i\in X \right \} _{i=1}^{\left|n\right|}$. The objective function here is the squared reconstruction error, defined as $L_{\theta}\left(x_{i} ; \hat{x}_{i}\right)=\frac{1}{n} \sum_{i=1}^{n}\left(\hat{x}_{i}-x_{i}\right)^{2}$.where $x_i$ is the original representation,and $\hat{x_i}$ is the reconstructed representation of the normal event. The error represents the semantic difference between the original and reconstructed events. Minimizing the reconstruction error of normal events allows the incremental discovery model to learn to capture and reconstruct the semantic features of normal events. Thus, the reconstruction error of normal events is lower than that of abnormal events, and the incremental discovery model can "recognize" normal events.

Finally, the decoder layer is re-bounded by minimizing the reconstruction error (lower part of Fig. 2), while the encoder layer includes potentially hidden representations that need to be frozen to prevent updating their parameters. Note that after the incremental discovery model learns to embed and aggregate normal events in the potential space $Z$ (middle of Fig. 2), the reason we retrain the decoder layer of the incremental discovery model (lower part of Fig. 2) is that the reconstruction error of abnormal events becomes easier to discriminate than normal events after the encoder layer learns to embed semantic features of normal events in the potential space $Z$.

\subsection{Incremental Event-type Mining Method based on Cluster Keyword Extraction}
The set of abnaomal events is obtained after the identification of self-encoder reconstruction errors by the incremental discovery model. Given that the subject of a document is usually expressed by keywords and can be visually reflected in the text representation vector obtained using text embedding models such as Bert and Doc2vec, the abnaomal events can be formed into clusters by clustering algorithms, and the names of the new event types represented by the clusters can be mined by extracting keywords from the clusters.

This paper evaluates the cohesive ability of real financial event data distributed in vector space by analyzing and comparing the k-means algorithm based on traditional vector distance partitioning, the density-based DBSCAN algorithm, the GMM Gaussian mixture model, the hierarchical clustering algorithm and the SOM network method; this paper adopts the GMM clustering results for the discovery of new event types.

Based on the generation of abnormal event clusters, this paper selects the LDA text topic model to extract the keywords of different abnormal event clusters and uses it as the reference set of event names for these clusters. The model consists of three levels of structure: word, topic, and document, where document-to-topic follows a polynomial probability distribution and topic-to-word follows a polynomial distribution. The similarity between the word distribution and the document distribution is calculated, and the words with the highest similarity are selected as the keywords for this document. The keyword generation process was divided into four steps:First, the topic distribution $\theta _i$ of the document is generated by sampling from the dirichlet distribution $\alpha$  of the document; Secondly, the topic $Z_{ij}$ of the $j$th word of the document is generated from the polynomial distribution $\theta _i$ of the topic; Dirichlet distribution from document $\beta$ that the word distribution $\varphi _{Z_{ij}}$ corresponding to topic $Z_{ij}$ is generated by sampling in; Finally, the word is generated by sampling from the polynomial distribution $\varphi _{Z_{ij}}$ of the word $\omega _{i,j}$. The name of the new event type is finally determined through the intervention of domain experts.

\section{Experimental Analysis}
\subsection{Data Preparation}
The financial domain is currently available for research use with the ChFinAnn built by Zheng et al\cite{r10}.DuEE-fin from the Baidu Thousand Words dataset, and the datasets for evaluation tasks in the financial domain released by Ali and Ant Group at the previous National Conference on Knowledge Graph and Semantic Computing (CCKS), whose goal of making the datasets available is to facilitate research on event extraction techniques in the financial domain. In recent years, the conference has released datasets for different information extraction tasks in the financial domain, such as the chapter-level event element extraction dataset for the financial domain, released in 2020, which includes 11 event types. The ChFinAnn dataset comes from ten years (2008-2018) of financial announcements of Chinese listed companies, including five event types with 32,000 data, which are significant events required to be disclosed by regulators; the DuEE-fin dataset is the latest chapter-level event extraction dataset in the financial sector released by Baidu, containing 11,700 chapters of 13 event types, News and announcements from the financial sector, covering many hard-to-solve problems in real-world application scenarios.
\begin{table}[]
\caption{Comparative presentation table of commonly used datasets for existing event extraction tasks}
\begin{tabular}{ccp{5cm}p{5cm}}
\hline
Dataset name & Number of event types & Event type                                                                                                                                                                                                   & Total number of event entries \\ \hline
DuEE-fin     & 13                    & Company Listing; Shareholder Reduction; Shareholder Increase; Corporate Acquisition; Corporate Financing; Share Repurchase; Pledge; Corporate Bankruptcy; Losses; Interviewed; Winning Bid; Executive Change & 1.17w                         \\
ChFinAnn     & 5                     & Shareholding freeze; Share repurchase; Shareholding reduction; Share capital increase; Shareholding pledge                                                                                                   & 3.2w                          \\
CCKS         & 11                    & Death of a senior executive; Corporate bankruptcy; Significant loss; Significant payout; Significant accident; Equity freeze; Equity pledge, increase in holdings; Reduction in holdings                     & ----                          \\ \hline
\end{tabular}
\end{table}

In this paper, DuEE-fin was selected as the base class data set, with 8608 events. The events of type "Equity Freeze" that did not exist in DuEE-fin in the ChFinAnn dataset were removed, and 9036 events were retrieved as the set of unknown normal events and pre-processed. Each unknown normal event was represented by a short text that could be assigned to an existing type in DuEE-fin.

To evaluate the ability of the model to identify abnormal events, this paper selects candidate financial events from the annual reports of companies listed on CCKS and the Shanghai and Shenzhen stock exchanges. The event entries belonging to known event types in the DuEE-fin dataset were removed from the synthetic set, and 300 events different from any event types in DuEE-fin were selected as the anomalous event set.
\subsection{Evaluation Methodology}
In this paper, the event induction performance of the incremental discovery model is evaluated on two datasets: events from DuEE-fin only and events from DuEE-fin+ (containing base class events and unknown events). As described in Section 3.1, the pairwise constraint matrix A is created using the existing events in DuEE-fin. The incremental discovery model is run 20 times independently and the average results are reported. For anomaly detection, the incremental discovery model is trained using only the events present in DuEE-fin, and its effectiveness is measured on the set of normal events in DuEE-fin, unknown normal events in DuEE-fin+, and anomalous events. This paper reports on the performance of the model in identifying events that trigger the same event type, using two methods to evaluate deep clustering techniques: the summed mean of purity and inverse purity (PIF) and the summed mean of BCubed precision and recall (BCF).

In this paper, the task of distinguishing anomalous events is defined as an anomaly detection task, using the area under the subject characteristic working curve (AUC ROC) and the area under the precision-recall curve (AUC PRC) as performance metrics. The anomaly scores generated due to the incremental discovery model and the baseline model are reconstruction errors and distances between event clusters or events, not binary labels. Without knowing the optimal anomaly threshold $\tau$ , AUC ROC and AUC PRC are ideal metrics that are threshold invariant and measure the quality of the model's prediction of abnormal events, regardless of the anomaly threshold $\tau$ \cite{r11}.

To evaluate the effectiveness of the new type of discovery model, at least ten iterations of each abnormal event clustering algorithm were conducted, and the hyperparameters of the model were adjusted using the clustering results obtained from each iteration, namely the ARI and AMI coefficients, and the best clustering results were selected as the final comparison basis for the clustering algorithm. This evaluation metric is used to test the accuracy of the clustering algorithm by calculating the probability that similar data in the sample set are correctly grouped without misclassifying different classes of data into one. We apply the winning model to the type discovery task in this paper (see Section 3.2 for details).
\subsection{Results and Analysis}
In terms of language models, this paper uses BERT and Doc2vec language models to generate contextual word embeddings to represent event texts, respectively. Both embedding methods can capture the semantic context and obtain better results in deep clustering.

Experiments with the known type data instance expansion method (Table 2) showed that the incremental discovery model was effective in assigning event types to events present or absent in DuEE-fin when the events were represented by a BERT embedding of their text. The high PIF and BCF scores indicate that the incremental discovery model generates homogeneous and complete clusters of events, with each cluster representing an event type since events sharing the same type, are grouped into a larger cluster rather than multiple smaller individual clusters. The better performance of the incremental discovery model comes from combining the prior knowledge of known events in DuEE-fin into pairs of information and representing the event text in a lower dimension, which makes it easier to distinguish the distance between two events with different types.

The experimental results show that the model tends to assign events to types with five or more existing events. One potential reason for this feature of the model is the dimensional catastrophe. In a high-dimensional vector space, the vectors of events within the same type are more widely distributed. Therefore, the more events a type contains, the more likely it is that a new event will be assigned to a type with a higher density.

\begin{table}[]
\caption{Clustering results of the clustering model and the two language models (LM) for events in the DuEE-fin and DuEE-fin \& ChinFinAnn datasets.}
\begin{tabular}{ccccccc}
\hline
\multirow{2}{*}{Model}                       & \multirow{2}{*}{Embed form}    & \multirow{2}{*}{LM} & \multicolumn{2}{c}{DuEE-fin} & \multicolumn{2}{c}{DuEE-fin \& ChinFinAnn} \\ \cline{4-7} 
                                             &                                &                     & PIF           & BCF          & PIF                  & BCF                 \\ \hline
\multirow{2}{*}{Incremental Discovery Model} & \multirow{2}{*}{Original text} & BERT                & 0.69525       & 0.54534      & 0.81642              & 0.70989             \\
                                             &                                & Doc2vec             & 0.20226       & 0.12049      & 0.35036              & 0.22434             \\ \hline
\end{tabular}
\end{table}

\begin{figure}
    \centering
    \includegraphics[width=17cm,height=7cm]{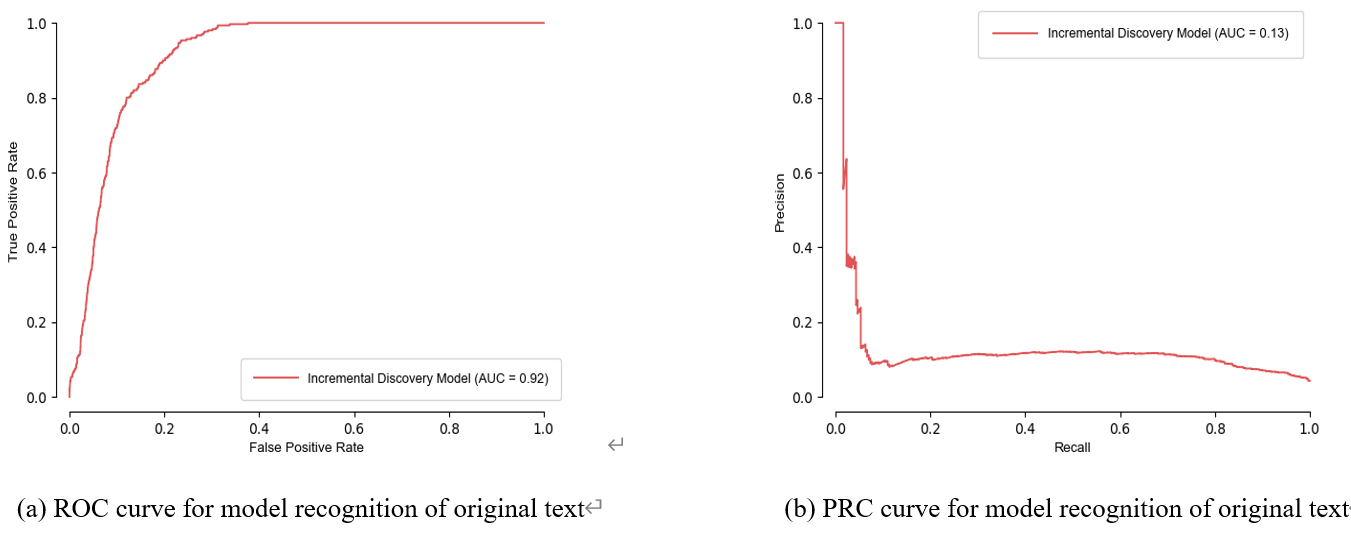}
    \caption{ROC curve (a) and PRC curve (b) of the model are used to identify events (abnormal events) in DuEE-fin that cannot be assigned any event type}
    \label{fig:my_label}
\end{figure}
The higher the AUC ROC and AUC PRC score in the abnormal event detection experiment, the better the model is at identifying abnormalities. Figure 4 shows that the incremental discovery model can effectively separate abnormal events from normal events, and the model can obtain an AUC ROC score of about 0.85 and an AUC PRC score of about 0.33, which indicates that the model can effectively distinguish abnormal events from normal events and also maintain some stability in the unbalanced data set.

Since there is no definite reconstruction error threshold $\tau$  for each screening. As shown in Figure 5, we used the reconstruction error value as the horizontal axis and the corresponding number as the vertical axis to plot the distribution of points, and observed that the reconstruction error of the model for unknown data contains a part of compact data including both normal and abnormal events, which we call the temporarily unprocessable event group. For this part, we take the approach of passing the temporarily unprocessable data around the peak (a mixture of normal and abnormal data, which cannot be distinguished directly) to the next round of processing in the new model and the new type library. The upper bound of the removed range is used as the minimum threshold for identifying abnormal events, and the lower bound of the removed range is used as the maximum threshold for identifying normal events (the minimum threshold is usually 0).

In the reconstruction error data set $\left\{R_{i}\right\}_{i=1}^{n}$ of the identified "abnormal event", the data set $\left\{R_{j}\right\}_{j=a}^{b}$ identified correctly is regarded as $TP$, and the data set $\left\{R_{j}\right\}_{j=c}^{d}$ identified incorrectly is regarded as $FP$, in which$j,a,b,c,d\in \left ( 1,n \right )$ , the accuracy rate $P=\frac{\sum_{j=a}^{b} R_{j}}{\sum_{i=1}^{n} R_{i}}$ is also calculated.
\begin{figure}
    \centering
    \includegraphics[width=10cm,height=7cm]{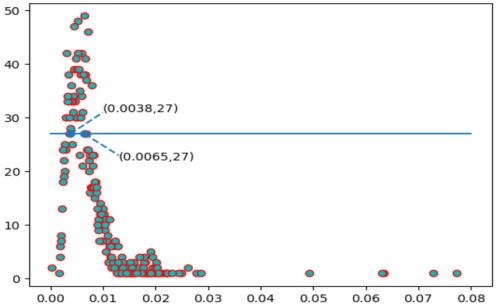}
    \caption{Error distribution diagram of abnormal event reconstruction}
    \label{fig:my_label}
\end{figure}
After determining the threshold of "normal" or "abnormal", we filtered out the abnormal data and passed it into the next process to get the new event type named. Experimentally, this method achieves an accuracy of 100\% for normal events and up to 70\% for abnormal events.

\begin{table}[]
\centering
\caption{Illustration of the results of different clustering methods applied to anomalous events}
\begin{tabular}{ccc}
\hline
\multirow{2}{*}{Model}  & \multicolumn{2}{c}{BERT embeds the original text} \\ \cline{2-3} 
                        & ARI                     & AMI                     \\ \hline
GMM                     & 0.731                   & 0.742                   \\
K-means                 & 0.585                   & 0.578                   \\
DBSCAN                  & 0.516                   & 0.623                   \\
Hierarchical clustering & 0.634                   & 0.650                   \\
SOM                     & 0.429                   & 0.469                   \\ \hline
\end{tabular}
\end{table}
The results of the evaluation indexes of the different clustering algorithms selected for the analysis of the new event-type incremental mining method are shown in Table 3, in which the GMM model performs better than the traditional k-means, DBSCAN, and hierarchical clustering based on vector space distance. This is because the experimental data of abnormal events are limited by the capability of the text embedding model used, and the segmentation boundaries between different class clusters are more complicated, which leads to a large number of misclassifications near the boundary lines of different class clusters by traditional clustering algorithms. However, the GMM Gaussian mixture model can fit arbitrarily shaped class cluster boundaries by multiple Gaussian distribution probability functions, and its way of determining the class of a sample by probability estimation can achieve better performance in dealing with the problem of ambiguous cluster boundaries.
\section{Related Work}
Deep clustering uses deep neural networks to learn feature representations that are beneficial for clustering tasks. When the dimensionality of the input feature space is very high, the similarity measures used by traditional clustering algorithms (such as k-means and hierarchical clustering methods) become unreliable, making direct clustering of the input embeddings ineffective\cite{r12}. Instead, deep clustering algorithms learn representations in a low-dimensional, cluster-friendly feature space. Xie et al.\cite{r13}proposed the Deep Embedding Clustering (DEC) algorithm, one of the most representative methods for deep clustering, which jointly learns feature representations and cluster assignments. To improve the clustering performance of DEC, Ren et al.\cite{r14}proposed a semi-supervised version of DEC (SDEC), which adds pairwise constraints to the feature learning process, allowing data points in the same cluster to become closer together and adjusting incorrect cluster assignments based on available data information. Zheng et al.\cite{r15}applied SDEC to the frame induction task to overcome the dimensionality disaster of high-dimensional contextual representation of lexical units. In this paper, anomaly detectors are introduced based on SDEC to accomplish the construction of features for different types of financial events, and thus accomplish the expansion of the number of instances for different event types.

Autoencoder is an unsupervised learning algorithm that learns to reconstruct its input using a deep neural network whose network structure consists of an encoder and a decoder. The encoder maps the original input vector to a lower-dimensional hidden representation, and the decoder maps the hidden representation back to the original input space. The difference between the original input vector and the reconstructed vector is called the reconstruction error. The auto-encoder learns to minimize this reconstruction error so that the auto-encoder approximates the recognition function\cite{r16}. After training the autoencoder to reconstruct anomaly-free data, the reconstruction error for anomalous data is very high\cite{r17} This makes it possible to detect anomalies. Autoencoder has been applied in the field of natural language processing to detect anomalies, such as network attacks\cite{r18}, and SMS spam\cite{r19}.

To recommend new types of naming for clusters of unusual events, the mainstream methods are based on statistics and feature engineering. The statistical-based methods are represented by TF-IDF and TextRank algorithms. TF-IDF mainly obtains the weight of words in a document through statistical information such as word frequency and determines the document keywords in order of priority.\cite{r20} TextRank is based on the idea of the PageRank algorithm, which uses the relationship between local words to build a co-occurrence network of candidate keywords.\cite{r21} The TextRank method is based on the PageRank algorithm, which uses the relationship between local words to construct a co-occurrence network of candidate keywords. The feature engineering-based methods include SVM, LDA, LSTM, and other traditional machine learning and deep learning methods. In these methods, the model is only as good as the features extracted.\cite{r22} The common idea is to use a word vector or text vector embedding model to map a text dictionary to a high-dimensional vector space based on its semantic similarity, and to complete the subject keyword extraction based on the text vector. Although the statistical-based approach is simple and easy to implement, it does not take into account the word order problem, and the selected keywords are mostly high-frequency common words, which do not provide a good overview of the semantics of the text. Therefore, in this task, we integrate the feature engineering-based extraction method.

\section{Conclusion}
In this paper, we propose a method to incrementally discover new event types from partially annotated financial event datasets. By conducting experiments on real datasets such as DuEE-fin, ChFinAnn, and CCKS, we discover new event types in the financial domain, such as the death of senior executives, corporate bankruptcy, major losses, major payouts, and major accidents, and validate the incremental discovery model of event types. The validity of the incremental event-type discovery model is verified. The work in this paper alleviates the shortcomings of the financial domain dataset, which is limited by scenarios and weak in reusability and can solve the current problem of the difficulty of using the massive and diverse new financial big data. In the future, we will further improve some of the shortcomings of this paper: for example, we can further propose a model that is more suitable for financial domain tasks in the process of anomalous event clustering and keyword extraction and naming; in identifying the features of different types of events, we will try to introduce graph neural networks to discover how the data of the same type will be consistent with each other.
% Your conclusion here
% \section*{Acknowledgments}
% This was was supported in part by......

%Bibliography
\bibliographystyle{unsrt}  
\bibliography{references}

\end{document}